\documentclass[a4paper,fleqn]{cas-sc}

\usepackage{natbib}
\usepackage{amssymb,amsmath}
\usepackage{graphicx}
\usepackage{hyperref}

\usepackage{subcaption}
\usepackage{booktabs}
\usepackage{tikz}
\usetikzlibrary{arrows.meta, positioning, shapes.geometric, fit, backgrounds}

\begin{document}
\let\WriteBookmarks\relax
\def\floatpagepagefraction{1}
\def\textpagefraction{.001}

\shorttitle{Large Models for BPHM}

\shortauthors{J. Liu et al.}

\title[mode = title]{Large Models for Battery Prognostics and Health Management: A Review and Future Roadmap}
\tnotemark[1]

\tnotetext[1]{This work was supported by a fellowship award from the Research Grants Council of the Hong Kong Special Administrative Region, China (Project No. CityU JRFS2526-1S09), and by the Research Grants Council of Hong Kong (Grants No. 11201023 and 11202224).}

\author[1]{Jiale Liu}[orcid=0009-0007-5881-3118]
\ead{Jiale.Liu@ed.ac.uk}

\author[2]{Huan Wang}[orcid=0000-0002-1403-5314]
\cormark[1]
\ead{hwan36@cityu.edu.hk}

\author[2]{Weicheng Wang}[orcid=0009-0008-9675-1293]

\author[2]{Rong Zhu}[orcid=0000-0003-4019-8587]

\author[2]{Qiqi Wang}[orcid=0000-0002-7721-1274]

\author[2]{Min Xie}[orcid=0000-0002-8500-8364]

\cortext[1]{Corresponding author}

\affiliation[1]{organization={School of Physics and Astronomy, The University of Edinburgh},
            city={Edinburgh},
            postcode={EH9 3FD},
            country={United Kingdom}}

\affiliation[2]{organization={Department of Systems Engineering, City University of Hong Kong},
            city={Hong Kong},
            country={China}}

\begin{abstract}
Battery Prognostics and Health Management (BPHM) is critical for ensuring the safe, reliable, and cost-effective operation of batteries across electric vehicles, grid storage, and consumer electronics. Conventional BPHM approaches, including physics-based models and task-centric deep learning methods, face fundamental challenges in computational efficiency and parameterization, cross-domain generalization, dependence on extensive labeled run-to-failure data, and model interpretability. The recent emergence of Large Models (LMs), built upon Transformer architectures and self-supervised pre-training, offers a transformative new paradigm to overcome these long-standing bottlenecks. This review provides the first comprehensive survey of LM applications in BPHM, systematically examining how these models address long-standing challenges in the field. We begin by elucidating the foundational technologies enabling LMs, including Transformer architectures, self-supervised learning, large-scale multimodal datasets, and parameter-efficient fine-tuning (PEFT) techniques. We then categorize recent progress in this area along four critical dimensions: mitigating data scarcity, enhancing generalization and robustness, integrating domain knowledge for improved interpretability, and enabling system-level automation. Despite promising initial results, significant challenges remain across data accessibility, intelligence validation, trustworthiness, and deployment feasibility. To guide future research, we propose a roadmap focused on building collaborative data ecosystems, validating intelligence for industrial applications, enhancing trustworthiness with physics-informed designs, and enabling efficient on-device deployment. This review establishes a systematic approach to understand and advance LM-driven BPHM, providing researchers and practitioners with essential insights for developing next-generation battery management systems capable of safe, reliable, and autonomous operation throughout battery lifecycles.
\end{abstract}


\maketitle

\section{Introduction}

As the world transitions towards sustainable energy, lithium-ion batteries have become a pivotal technology powering this transformation, with applications spanning electric vehicles (EVs) \citep{zeng_commercialization_2019}, grid-scale energy storage systems \citep{chen_applications_2020}, and consumer electronics \citep{lu_research_2019, kong_strategies_2021, wang_reviving_2018}. However, the degradation of battery performance presents significant economic and safety challenges \citep{wang_early_2024, han_review_2019}, making accurate Battery Prognostics and Health Management (BPHM) critical for ensuring safe, reliable, and long-life operation \citep{meng_review_2019}. Meanwhile, the rapid emergence of large-scale models is fundamentally reshaping scientific and engineering disciplines. This paradigm shift raises a compelling question: how can BPHM benefit from the representational power, generalization capabilities, and scalability of large models to address longstanding challenges in battery state estimation and lifetime prediction?

Early research on BPHM relied heavily on physics-based models that attempted to encapsulate the complex internal electrochemical processes of a battery within mathematical frameworks \citep{brosa_planella_continuum_2022}. These models aim to simulate phenomena such as lithium-ion diffusion, ionic transport, interfacial reaction kinetics, and solid electrolyte interphase (SEI) growth \citep{wang_physics-informed_2025, deng_reduced-order_2021}. Prominent examples include the Doyle-Fuller-Newman (DFN) model \citep{brosa_planella_continuum_2022, tredenick_multilayer_2024}, often considered the gold standard for its detailed electrochemical fidelity, and its simplified variant, the Single Particle Model (SPM) \citep{hassanaly_pinn_2024}. The primary advantage of this paradigm lies in its high degree of interpretability, as each model parameter corresponds to a distinct physical property, theoretically enabling strong extrapolation capabilities to unseen operating conditions. However, physics-based models are beset by two critical drawbacks: immense computational complexity, arising from the need to solve coupled partial differential equations, which renders them unsuitable for real-time applications in a BMS; and the significant challenge of parameterizing the model, which requires numerous electrochemical parameters that are difficult to measure directly and vary with cell age and temperature \citep{brosa_planella_continuum_2022, ali_comparison_2024}.

This gap led to the adoption of data-driven methods, specifically machine learning (ML), where models learn directly from data. This initial phase utilized algorithms such as Support Vector Machines (SVMs), which map input features to a high-dimensional space to find an optimal separating hyperplane for regression or classification \citep{feng_online_2019, klass_support_2014, patil_novel_2015}, and ensemble methods like Random Forests \citep{mawonou_state--health_2021, li_random_2018, wang_optimized_2023, gotz_random_2024}, which aggregate predictions from multiple decision trees to enhance robustness. While effective, the performance of these ML models is critically contingent on a process of manual feature engineering, where domain experts should design and extract salient indicators of degradation from raw signals, such as the integral of specific voltage ranges or the duration of constant-current charging phases \citep{patil_novel_2015}. This reliance on handcrafted features constitutes a significant bottleneck, limiting both the scalability and adaptability of the models.

The proliferation of deep learning (DL) subsequently catalyzed a paradigm shift towards what can be called the ``conventional DL-based methods,'' driven by the promise of automated, end-to-end feature extraction. A diverse array of architectures was explored to this end. Convolutional Neural Networks (CNNs), adapted from their success in image analysis, apply 1D convolutional filters to battery time-series data. This approach allows them to autonomously learn and identify local morphological patterns, such as shifts in the voltage curve's ``knee point,'' which are highly indicative of degradation \citep{costa_li-ion_2022, lee_convolutional_2023, wang_fesd_2025}. To better capture the cumulative and time-dependent nature of battery aging, the field also widely adoptes Recurrent Neural Networks (RNNs) \citep{zhao_rnn_2023, feng_state--charge_2021}, particularly their gated variants, Long Short-Term Memory (LSTM) \citep{chen_soc_2023, chai_novel_2024} and Gated Recurrent Units (GRU) \citep{jiao_gru-rnn_2020, yang_state--charge_2019, yao_surface_2023}. These models process data sequentially, employing internal memory states to model temporal relationships across charging and discharging cycles. Other architectures, such as Autoencoders, were also utilized, primarily for unsupervised tasks like anomaly detection by learning a compressed representation of normal operational data \citep{zhang_novel_2023, li_fault_2025, sun_anomaly_2023}.

\begin{figure}[pos=htbp]
    \centering
    \includegraphics[width=\textwidth]{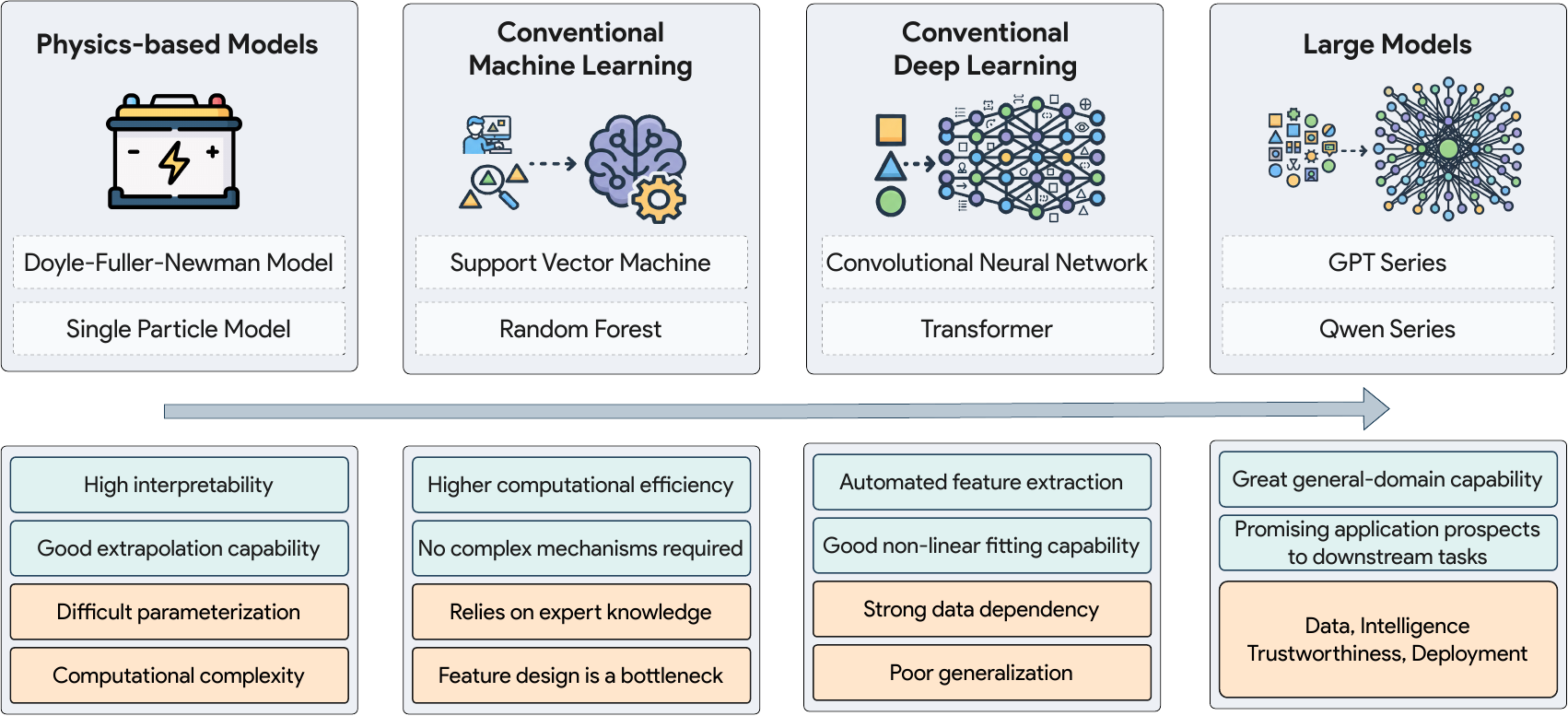}
    \caption{An overview of the evolution of modeling methods in BPHM.}
    \label{fig:model_evolution}
\end{figure}

More recently, the Transformer architecture emerged as a powerful alternative \citep{sun_state--health_2025, wang_online_2022, zhao_battery_2023, chen_lithium-ion_2025}. Its core innovation, the self-attention mechanism, enables the model to dynamically weigh the influence of all past time steps when analyzing a current state, thereby overcoming the long-range dependency challenges inherent in RNNs and providing a superior capability to model degradation over extended operational lifetimes \citep{gomez_li-ion_2024, NIPS2017_3f5ee243}. The progression of these modeling methods, along with their respective strengths and weaknesses, is visually summarized in \autoref{fig:model_evolution}.

However, despite the increasing complexity of the architecture, all these models are still deployed within the same constrained framework. They are typically trained from scratch in a supervised manner on specific, often limited, datasets to perform a singular objective, which can be characterized as Task-Centric Model Development. While this approach has yielded notable successes and significantly advanced the intellectualization of BPHM, it is bound by fundamental limitations. First, it relies heavily on the expensive and time-consuming run-to-failure labeled data on batteries \citep{lu_deep_2023}. Second, its task-centric nature precludes knowledge sharing across different but related BPHM tasks and objectives \citep{bao_multi-state_2023}. Finally, and most critically, the resulting models suffer from poor generalization, as their performance often deteriorates when confronted with different battery chemistries or operational conditions \citep{chen_deep-learning-based_2024}. These collective bottlenecks signify that the conventional deep learning paradigm, irrespective of the underlying model architecture, is reaching its functional limits, underscoring the urgent need for a new theoretical and practical framework.

\begin{figure}[pos=htbp]
    \centering
    \includegraphics[width=\textwidth]{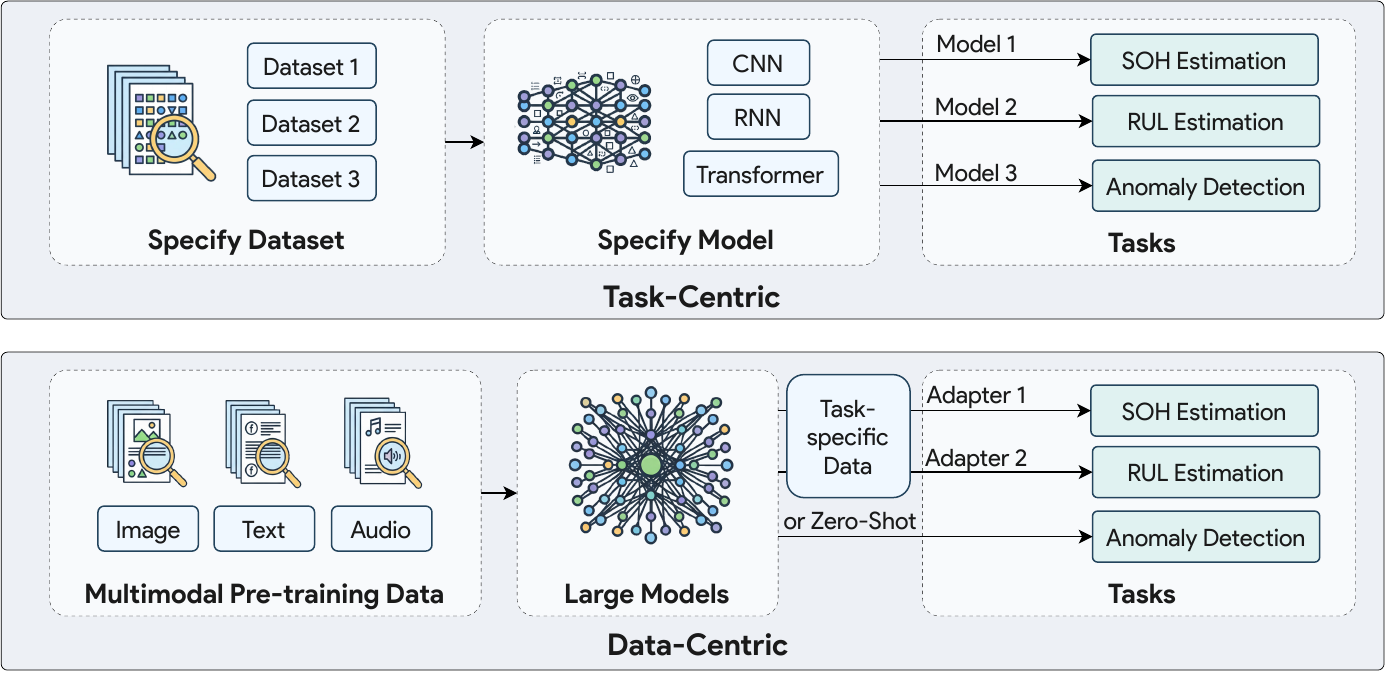}
    \caption{A comparison between the Task-Centric paradigm and the LM-based Data-Centric paradigm in BPHM. The Task-Centric approach requires training separate, specialized models for each task, while the Data-Centric approach leverages a single pre-trained large model that can be adapted to multiple downstream tasks.}
    \label{fig:paradigm_shift}
\end{figure}

Against this backdrop, Large Models (LMs) represent a departure from the task-specific supervised training that defines conventional BPHM. An LM is, at its core, a deep learning model with massive parameterization---typically hundreds of millions to hundreds of billions of parameters---trained on large-scale corpora via self-supervised learning. The term covers several distinct types: \textit{Large Language Models} (LLMs), such as the GPT series \citep{brown_language_2020, openai_gpt-4_2024}, LLaMA series \citep{touvron_llama_2023, grattafiori_llama_2024}, and Qwen series \citep{bai_qwen25-vl_2025, yang_qwen3_2025}, are pre-trained on text for natural language understanding, generation, and reasoning; \textit{Large Vision Models}, such as the Segment Anything Model (SAM) \citep{Kirillov_2023_ICCV, ravi_sam_2024}, handle visual perception; \textit{Large Multimodal Models}, such as CLIP \citep{radford_learning_2021}, learn joint representations across modalities; and \textit{Time-Series Foundation Models}, such as TimeGPT \citep{garza_timegpt-1_2024} and Lag-Llama \citep{rasulLagLlamaFoundationModels2024}, extend this pre-training logic to temporal signals. Despite their differences, all four types rest on the same foundation: the Transformer architecture, self-supervised pre-training, and task-agnostic adaptation. That shared design allows a single pre-trained model to be reused across many downstream tasks, marking a shift from task-centric model development to a data-centric paradigm (see \autoref{fig:paradigm_shift}). What sets LMs apart from conventional data-driven BPHM is not merely scale: they introduce capabilities that have no real equivalent in prior methods----(1) emergent reasoning via Chain-of-Thought (CoT) prompting and in-context learning, enabling multi-step diagnostic inference without task-specific training; (2) agentic behavior, where models plan, retrieve information, and call external tools autonomously; and (3) a pre-training and adaptation paradigm that decouples knowledge acquisition from task execution, letting one model serve many applications through fine-tuning or prompting.

This shift is fundamentally enabled by three defining characteristics of LMs. First, almost all of them are universally built upon the Transformer architecture, which offers a powerful mechanism for capturing long-range dependencies in sequential data, especially the cumulative, non-linear aging trajectories of batteries over extended lifecycles. \citep{kotei_systematic_2023}. Second, their knowledge is acquired through self-supervised pre-training on vast, often web-scale corpora rather than narrow, task-specific datasets, which allows models to learn robust general representations, thereby circumventing the industry's reliance on scarce and expensive run-to-failure labeled battery data \citep{kotei_systematic_2023, shen_efficient_2025}. Third, they are designed as reusable, general-purpose infrastructures, and downstream tasks are addressed through fine-tuning or prompting rather than training task-specific networks from scratch \citep{zheng_learning_2025}. This flexibility provides a distinct advantage in handling the diverse chemistries and operating conditions typical in BPHM, enabling cross-domain generalization that traditional models struggle to achieve.

These advancements are driving researchers to explore how LMs can address the long-standing challenges in BPHM across tasks such as SOH estimation, RUL prediction, and anomaly detection. Encouragingly, early but promising research outcomes have emerged in multiple pathways, including adapting LLMs for time-series analysis through textualization and prompt engineering, fine-tuning general-purpose time-series foundation models on battery-specific datasets, and developing physics-informed adaptation techniques to enhance model robustness and reliability. These efforts will be detailed in later sections.

However, despite these promising initial explorations, the application of LMs to BPHM is still in its infancy. The field currently lacks a systematic review to unify these disparate efforts, thoroughly analyze the underlying principles, and illustrate the current status, challenges, and future prospects. To bridge this gap, this paper aims to conduct a thorough and systematic survey of this emerging domain, providing a comprehensive overview and a clear roadmap for future research. Specifically, the main contributions of this review are summarized as follows:

\begin{itemize}
    \item This review provides a comprehensive overview of the key enabling technologies for LMs in BPHM, systematically answering what constitutes this new paradigm and its core components.
    \item Drawing from the current state-of-the-art, this paper systematically elucidates why LMs are a compelling solution for BPHM and how they can be effectively built and adapted for battery-specific applications.
    \item This paper outlines a detailed research roadmap for LM-based BPHM, offering a structured understanding of the critical challenges, future research directions, and potential opportunities for advancement.
    \item To the best of our knowledge, this is the first systematic review dedicated to the application of LMs in BPHM, and it is expected to provide valuable guidance for researchers and practitioners in this rapidly evolving field.
\end{itemize}

The rest of this paper is organized as follows: \autoref{sec:enablers} reviews the core technologies that are fundamental to LMs. \autoref{sec:progress} surveys the recent progress and different pathways to apply LMs for BPHM. \autoref{sec:challenges_sec} identifies and discusses the current challenges in this emerging field. \autoref{sec:roadmap_sec} outlines a future research roadmap with potential solutions, and \autoref{sec:conclusions} concludes the paper.

\section{Key Enablers of Large Models for Battery Intelligence}
\label{sec:enablers}

The emergence of LMs represents a convergence of technological advancements that collectively provide the necessary ingredients for building, training, and adapting models at a scale previously unattainable in the energy sector. While these technologies originated in natural language processing and computer vision, their underlying principles offer direct solutions to the most persistent challenges in BPHM. This section is organized around these challenges: \autoref{sec:transformer} introduces the Transformer and self-attention mechanism, \autoref{sec:ssl} covers self-supervised learning, \autoref{sec:multimodal} discusses multi-modal data fusion, and \autoref{sec:adaptation} presents parameter-efficient fine-tuning strategies.

\subsection{Transformer and Self-Attention Mechanism}
\label{sec:transformer}
A defining characteristic of battery degradation is its cumulative, long-range temporal nature: the health state at any given cycle is shaped by the superposition of electrochemical stress events spanning hundreds to thousands of preceding cycles, including subtle shifts in solid-electrolyte interphase growth, lithium plating nucleation, and active material loss \citep{birkl_degradation_2017}. Conventional recurrent architectures (LSTMs, GRUs) process data sequentially and suffer from vanishing gradients over such extended horizons, making it difficult to relate early-life abuse events to late-life capacity knee points \citep{karita_comparative_2019, wen_rnns_2024}. The Transformer architecture, introduced by Vaswani et al. \citep{NIPS2017_3f5ee243}, directly addresses this pain point. By replacing sequential processing with a self-attention mechanism that models relationships between all elements in a sequence regardless of their temporal distance \citep{NIPS2017_3f5ee243, soydaner_attention_2022}, Transformers can capture the long-range dependencies inherent in battery degradation data in a fully parallelizable manner.

\begin{figure}[pos=htbp]
    \centering
    \begin{subfigure}[t]{0.5\textwidth}
        \centering
        \includegraphics[height=6cm]{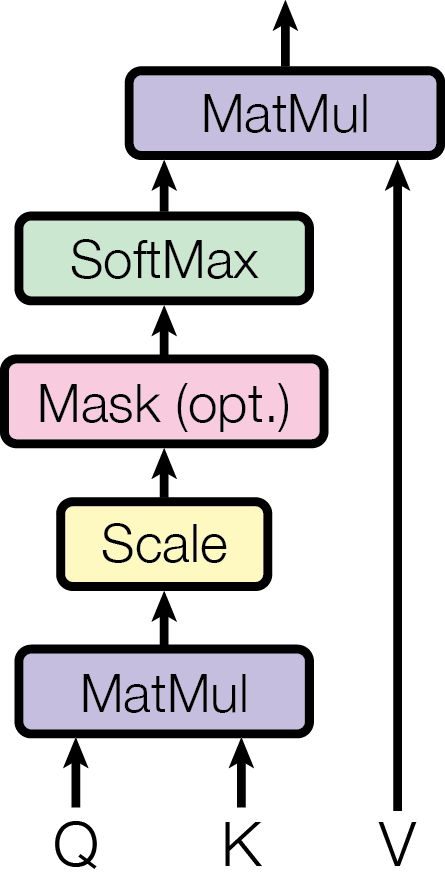}
        \caption{Scaled Dot-Product Attention}
    \end{subfigure}%
    ~
    \begin{subfigure}[t]{0.5\textwidth}
        \centering
        \includegraphics[height=6cm]{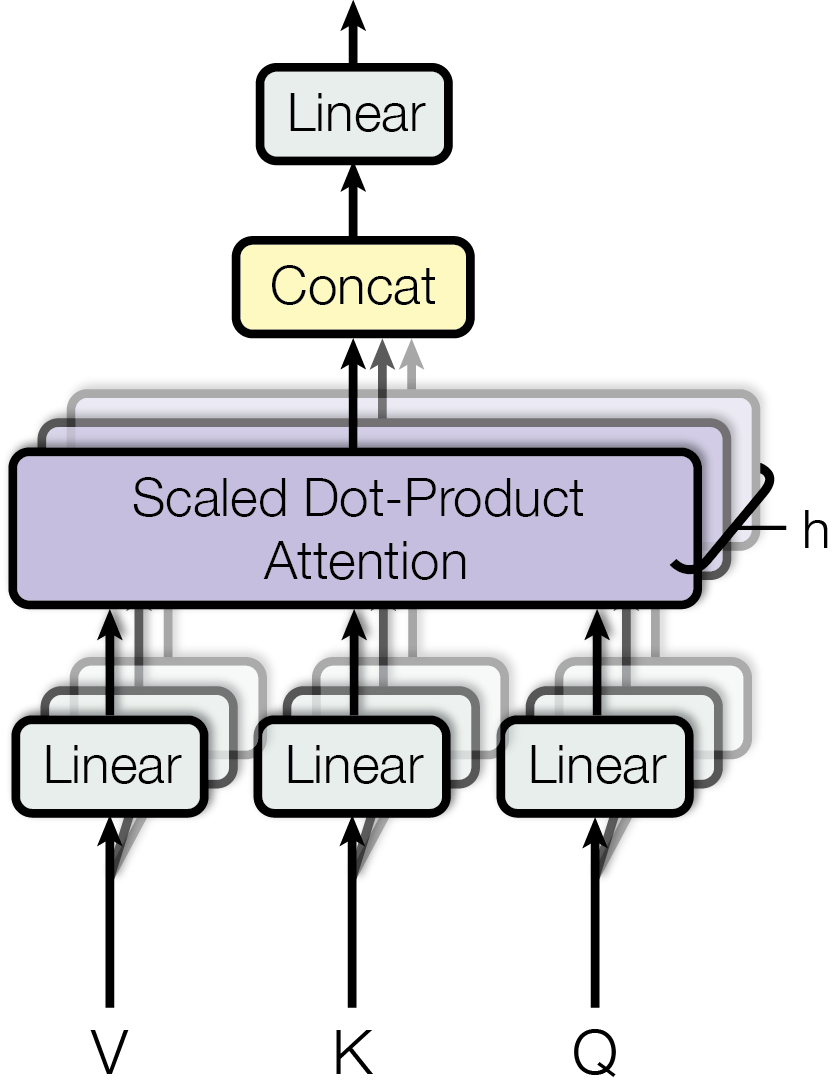}
        \caption{Multi-Head Attention}
    \end{subfigure}
    \caption{The architecture of the self-attention mechanism in Transformers~\citep{NIPS2017_3f5ee243}. (a) The core Scaled Dot-Product Attention computes attention scores using Query (Q), Key (K), and Value (V) vectors. (b) The Multi-Head Attention mechanism runs multiple scaled dot-product attention operations in parallel and concatenates their outputs, allowing the model to jointly attend to information from different representation subspaces.}
        \label{fig:attention}
\end{figure}

At its core, the self-attention mechanism enables a model to weigh the relevance of different historical time steps when estimating the current battery state. Each input embedding—representing a snapshot of voltage, current, or temperature—is transformed into Query ($Q$), Key ($K$), and Value ($V$) vectors. The attention score is calculated by the scaled dot product of the Query with all Keys, determining which historical segments contain the most salient information regarding degradation patterns. This mechanism allows the model to dynamically focus on critical aging indicators, such as the shift in voltage plateaus or the onset of the knee-point, effectively ignoring irrelevant noise in the charging curve. To capture diverse aging features simultaneously, such as capacity fade and resistance increase, Transformers employ multi-head attention to attend to different representation subspaces in parallel. The architectural details are illustrated in \autoref{fig:attention}.

However, a critical bottleneck for applying standard self-attention to high-frequency battery data is its quadratic complexity, $O(n^2)$, with respect to sequence length $n$ \citep{keles_computational_2022, roy_efficient_2021}. A standard charging profile sampled at 1Hz can generate thousands of points, making full attention computationally prohibitive. To address this, efficient attention mechanisms relevant to time-series have been developed. Sparse attention methods reduce complexity by computing scores for only a subset of time steps, akin to focusing on key operational windows \citep{roy_efficient_2021, huang_sparse_2024}. Linear attention methods approximate the attention matrix to achieve $O(n)$ complexity, facilitating the processing of extremely long cycling lifetimes \citep{guo_beyond_2023, han_flatten_2023}. Additionally, hardware-aware optimizations like FlashAttention \citep{dao_flashattention_2022} maximize memory throughput, enabling the training of Foundation Models on extensive historical battery databases.

\subsection{Self-Supervised Learning for Signal Representation}
\label{sec:ssl}

Modern BMS platforms generate terabytes of unlabeled operational data, yet ground-truth labels---such as capacity fade from reference performance tests or verified failure-mode annotations---remain expensive to obtain \citep{lu_deep_2023}. Self-Supervised Learning (SSL) directly addresses this imbalance by leveraging the abundant unlabeled field data as its own supervision signal. By devising ``pretext tasks'' where the data itself provides the training objective \citep{zhang_self-supervised_2024}, SSL enables models to learn high-level representations of electrochemical dynamics without human annotation, creating a general-purpose feature extractor that can subsequently be fine-tuned for downstream tasks with only minimal labeled samples \citep{ding_parameter-efficient_2023}.

The primary SSL methodologies have direct analogs in battery signal processing. \textit{Masked Modeling}, or Denoising, learns representations by reconstructing masked portions of the input \citep{chen_masked_2023, hondru_masked_2025}. In BPHM, this is particularly valuable for handling sensor faults and data transmission loss. By training a model to reconstruct masked segments of a voltage or current profile, the network learns the intrinsic correlations of the charging curve. This capability can be directly applied to recover missing sensor data or to perform anomaly detection, where the inability to accurately reconstruct a segment indicates a potential defect or safety hazard. This mirrors the success of BERT in language \citep{devlin_bert_2019} and Masked Autoencoders in vision \citep{he_masked_2021}.

\textit{Autoregressive (AR) Modeling} trains the model to predict the next token in a sequence based on preceding context \citep{NEURIPS2019_dc6a7e65}. For batteries, this translates to forecasting future voltage responses or capacity trajectories based on historical operation windows. This approach is fundamental for online state estimation and short-term safety monitoring. Finally, \textit{Contrastive Learning} trains models to differentiate between similar and dissimilar samples \citep{le-khac_contrastive_2020}. In the battery domain, this involves maximizing the similarity between different augmented views of the same charging cycle (e.g., with added noise or time-warping) while separating it from cycles of different batteries or aging states. This forces the model to learn robust representations of the State of Health (SOH) that are invariant to temporary ambient fluctuations, similar to the principles used in SimCLR \citep{chen_simple_2020} and CLIP \citep{radford_learning_2021}.

\subsection{Large-Scale and Multimodal Data Fusion}
\label{sec:multimodal}

Battery health diagnostics is inherently a multi-modal and multi-scale problem that conventional single-modality models struggle to address holistically. A comprehensive understanding of degradation requires integrating heterogeneous data sources that span different physical domains: 1D time-series signals (voltage, current, temperature profiles during cycling), 2D spectral and visual data (Electrochemical Impedance Spectroscopy (EIS) Nyquist plots, thermal images, and post-mortem microstructure micrographs), and unstructured textual data (cell datasheets, maintenance logs, and published research) \citep{yin_survey_2024}. Yet traditional deep learning pipelines typically process each data stream in isolation, losing the diagnostic synergy that arises from cross-modal corroboration.

Addressing this challenge also depends on data scale, adhering to scaling laws where model performance improves with data volume \citep{kaplan_scaling_2020}. This has prompted a shift from small, laboratory-generated datasets to large-scale repositories spanning diverse chemistries and operational profiles, enabling models to acquire a ``general sense'' of electrochemical behavior and unlock zero-shot generalization to unseen cell types. LMs are uniquely positioned for multi-modal fusion by projecting disparate modalities into a shared embedding space \citep{zong_self-supervised_2025}. For instance, a multi-modal LM could associate a textual description of a ``high-nickel cathode'' with features in the voltage relaxation curve and visual patterns in an EIS Nyquist plot, enabling corroboration across channels---detecting a fault via thermal anomalies even if voltage sensors remain nominal \citep{radford_learning_2021, qi_large_2023}.

\subsection{Effective Adaptation}
\label{sec:adaptation}

The battery industry encompasses a vast diversity of cell chemistries (LFP, NMC, NCA, LCO, solid-state), form factors (cylindrical, pouch, prismatic), and deployment conditions (tropical EV fleets, cold-climate grid storage, consumer electronics). Each combination exhibits distinct degradation signatures, meaning that a model trained on one chemistry--condition pair may perform poorly on another. Retraining a full large model for every new cell variant is computationally prohibitive and defeats the purpose of building general-purpose foundation models. This creates a critical need for lightweight adaptation strategies that can efficiently specialize a pre-trained LM to a target battery domain while preserving its broad learned representations. One approach is \textit{Prompting}, which involves transforming battery data into a format interpretable by the model, often by treating signal patches as tokens or by embedding numerical data alongside textual instructions \citep{li_practical_2023}. This allows a frozen LLM to perform time-series forecasting or anomaly detection simply by being ``prompted'' with historical data and a task description, leveraging its reasoning capabilities without weight updates \citep{agarwal_many-shot_2024}.

\begin{figure}[pos=htbp]
    \centering
    \includegraphics[width=\textwidth]{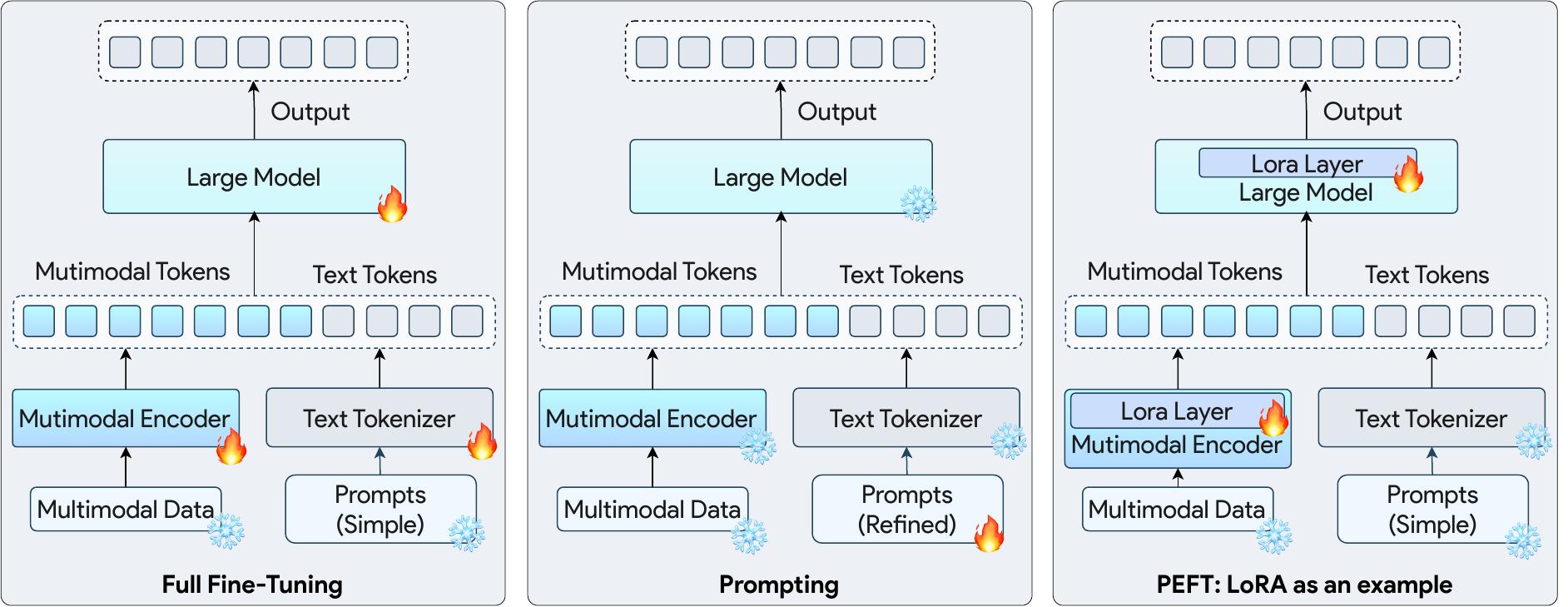}
    \caption{An illustration of different strategies for adapting large models. (Left) Full fine-tuning updates all model parameters, which is computationally expensive. (Center) Prompting keeps the model frozen and relies on refining input prompts to guide behavior. (Right) PEFT, exemplified by LoRA, freezes the base model and trains only a small number of supplementary parameters (LoRA layers), offering a computationally efficient compromise. The fire icon indicates trainable components, while the snow icon indicates frozen components.}
    \label{fig:adaptation_strategies}
\end{figure}

The second approach, \textit{Parameter-Efficient Fine-Tuning (PEFT)}, addresses the computational prohibitive cost of retraining massive models for every new battery model \citep{han_parameter-efficient_2024}. Instead of updating all parameters, PEFT techniques like Low-Rank Adaptation (LoRA) freeze the pre-trained backbone and inject a small number of trainable rank-decomposition matrices \citep{hu_lora_2021}. This is particularly beneficial for the scalability of BPHM, as it only requires training lightweight adaptation modules for each specific vehicle model or battery manufacturer to deploy the LM across a range of diverse vehicles. Other techniques like Adapters \citep{hu_llm-adapters_2023} and Prefix-Tuning \citep{li_prefix-tuning_2021} similarly allow for the injection of task-specific knowledge while preserving the model's general robustness. A conceptual comparison of these strategies is provided in \autoref{fig:adaptation_strategies}.

\section{Progress in Large-Model-driven BPHM}
\label{sec:progress}

LM-driven BPHM fundamentally diverges from conventional deep learning by redefining battery state estimation through universal representation learning and semantic reasoning. Rather than designing task-specific architectures trained from scratch, the focus shifts to grounding general-purpose capabilities, such as few-shot adaptation, cross-modal transfer, and generative reasoning, in the battery domain.

\begin{figure}[pos=htbp]
    \centering
    \includegraphics[width=0.6\linewidth]{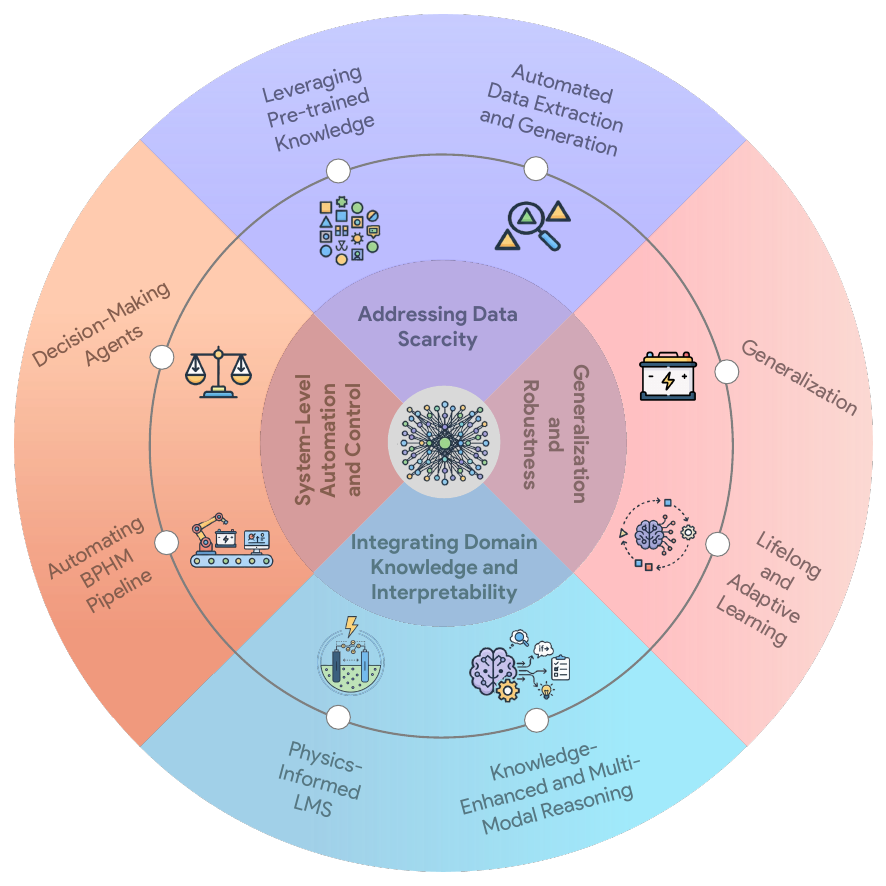}
    \caption{An overview of recent progress in applying LMs for BPHM.}
    \label{fig:lm_progress}
\end{figure}

To provide a structured analysis of this field, we categorize recent contributions according to the specific problem they are designed to address. As illustrated in \autoref{fig:lm_progress}, the progress in LM-driven BPHM is systematically reviewed along four critical dimensions: (1) addressing domain data scarcity, (2) enhancing model generalization and robustness, (3) integrating domain knowledge and improving interpretability, and (4) enabling system-level automation and control.

\subsection{Addressing Domain Data Scarcity}

Given the prohibitive cost of generating run-to-failure datasets, LMs offer multiple strategies to reduce the labeled data requirement that has constrained conventional supervised models.

\subsubsection{Leveraging Pre-trained Knowledge}

The primary strategy to mitigate data scarcity is to exploit the internal representations acquired during pre-training, which significantly reduces the sample requirement for downstream tasks. Fan et al. \citep{fan_analyzing_2025} demonstrated that by distilling a LLM into a smaller predictor, the system could identify degradation patterns using a fraction of the data required by varying RNNs. This efficiency is further validated by Peng et al. \citep{peng_large-language-model-enabled_2025} in their ``Internet of Batteries'' framework. By employing a foundation model (Lag-Llama) pre-trained on diverse time-series corpora, they achieved state-of-the-art prediction accuracy using only roughly 0.72\% of the target dataset for fine-tuning. Similarly, in anomaly detection, Cheng et al. \citep{cheng_batterygpt_2024} utilized the BatteryGPT framework to perform few-shot classification. By leveraging pre-aligned visual and textual encoders, the model could generalize to unseen defect types without requiring the extensive library of fault examples necessary for conventional supervised classifiers.

These results suggest that latent representations learned from vast pre-training corpora are surprisingly transferable to electrochemical signals, shifting the focus from training feature extractors from scratch to aligning pre-learned representations with battery dynamics.

\subsubsection{Automated Data Extraction and Generation}

Beyond efficient adaptation, LMs are utilized to actively expand the available data ecosystem by extracting latent knowledge from literature. To unlock unstructured data, Lee et al. \citep{lee_data-driven_2024} developed an automated pipeline that employs an LLM as a semantic parser. The model reads scientific papers to extract cell specifications and pairs them with digitized cycling curves, successfully structuring a database of over 8,000 cells from raw PDFs. Complementing extraction with generation, Huang et al. \citep{huang_datagen_2024} introduced DataGen, a framework where the LM acts as a controllable generator. By conditioning the model on specific attributes, it synthesizes high-fidelity battery data that adheres to physical consistency checks, effectively solving the ``cold start'' problem for data-starved applications.

By converting unstructured literature into structured databases, LMs enable BPHM systems to self-enrich their training environments, reducing reliance on expensive physical experimentation.

\subsubsection{Landscape of Open-Source Battery Datasets and Benchmarks}

Training LMs at scale demands large, diverse, and well-curated datasets. To meet this need, the battery research community has also made significant efforts in assembling and openly releasing cycling datasets that span a wide range of chemistries, operating conditions, and data modalities. \autoref{tab:datasets} provides a structured overview of these major publicly available resources.

\begin{table}[pos=htbp]
\centering
\caption{Summary of major open-source battery datasets for BPHM research. V = Voltage, I = Current, T = Temperature, EIS = Electrochemical Impedance Spectroscopy, Cap. = Capacity.}
\label{tab:datasets}
\scriptsize
\begin{tabular}{p{3.0cm}p{2.3cm}p{0.9cm}p{1.1cm}p{4.5cm}p{1.4cm}}
\toprule
\textbf{Dataset} & \textbf{Chemistry} & \textbf{\#Cells} & \textbf{Type} & \textbf{Cycling Conditions} & \textbf{Data Types} \\
\midrule
NASA PCoE \citep{PrognosticsCenterExcellence} & LCO & 34 & 18650 & Charge/discharge/EIS at 4--43$^\circ$C & V, I, T, EIS \\
CALCE \citep{BatteryResearchData} & LCO, LFP, NMC & Multiple & Mixed & CC-CV, DST, FUDS, US06, OCV, EIS & V, I, T, Cap., EIS \\
MIT-Stanford \citep{seversonDatadrivenPredictionBattery2019} & LFP & 124 & 18650 & Fast-charging (varying C-rates), 30$^\circ$C & V, I, T, Q(V) \\
Stanford \citep{attiaClosedloopOptimizationFastcharging2020} & LFP & 192 & 18650 & 224 six-step fast-charging protocols & V, I, T, Cap. \\
Oxford \citep{howeyOxfordBatteryDegradation2017} & LCO & 8 & Pouch & Artemis drive cycle, 40$^\circ$C & V, I, T, Cap. \\
RWTH Aachen \citep{liOneshotBatteryDegradation2021} & NMC & 48 & 18650 & Uniform protocol, 25$^\circ$C & V, I, T \\
SNL (Sandia) \citep{pregerDegradationCommercialLithiumIon2020} & NCA, NMC, LFP & $\sim$60 & 18650 & Varied temp., DOD, discharge rate & V, I, T, Cap. \\
HNEI (Hawaii) \citep{devieIntrinsicVariabilityDegradation2018} & NMC-LCO & 15 & 18650 & C/2 charge, 1.5C discharge, 25$^\circ$C & V, I, T, Cap. \\
XJTU \citep{wangOpenAccessDataset2024} & NMC & 55 & 18650 & 6 charge/discharge strategies & V, I, T \\
HUST \citep{maRealtimePersonalizedHealth2022} & LFP & 77 & 18650 & Identical fast-charge, 77 discharge protocols, 30$^\circ$C & V, I, T, Cap. \\
Tongji \citep{zhuDatadrivenCapacityEstimation2022} & 3 systems & 130 & 18650 & 3 temps, 6 protocols & V, I, T, Cap. \\
ISU-ILCC \citep{liPredictingBatteryLifetime2024} & NMC/graphite & 238 & 502030 pouch & 63 conditions (rate, DOD), 30$^\circ$C & V, I, T, Cap. \\
\bottomrule
\end{tabular}
\end{table}

Several observations emerge from this landscape. The vast majority of publicly available datasets are generated under controlled laboratory conditions with small-format cells (predominantly 18650 cylindrical), leaving real-world deployment scenarios---pack-level field data from EVs, grid storage, and charging infrastructure---severely underrepresented. The datasets also vary substantially in scale (from 8 to 238 cells), chemistry coverage, and cycling protocols, making cross-dataset comparisons difficult without careful harmonization.

To address this fragmentation, recent efforts have focused on unified benchmarking platforms. BatteryML \citep{zhangBatteryMLOpensourcePlatform2024} provides an open-source platform unifying data preprocessing, feature extraction, and model training across seven datasets. BatteryLife \citep{tanBatteryLifeComprehensiveDataset2025} represents the most comprehensive benchmark to date, integrating 16 source datasets encompassing 990 batteries across 59 chemical systems and 8 cell formats---including, for the first time, zinc-ion, sodium-ion, and industry-scale large-capacity lithium-ion cells---with 18 benchmark methods spanning MLPs, Transformers, CNNs, and RNNs. BatteryArchive.org serves as an open repository with visualization and analysis capabilities. Despite these advances, critical gaps remain: most benchmarks focus on cell-level laboratory data, and standardized evaluation of emerging LM-based approaches against classical baselines is still lacking.

\subsection{Enhancing Generalization and Robustness}

A long-standing challenge in BPHM is the poor generalization of traditional data-driven models, whose performance degrades when deployed to unseen chemistries or operating conditions. LMs, by leveraging knowledge from large-scale pre-training, provide a generalizable foundation that can be efficiently adapted to diverse battery scenarios.

\subsubsection{Cross-Domain and Cross-Chemistry Generalization}

A dominant approach to achieving generalization is to decouple the core reasoning mechanism from specific data formats using prompting and foundation models. Bian et al. \citep{bian_exploring_2024, bian_hybrid_2025} and Qiu et al. \citep{qiu_prompt-driven_2024} pioneered prompt-learning strategies where numerical battery data is reformatted into textual or tokenized prompts. This abstraction allows a single LM to estimate SOC \citep{bian_exploring_2024, bian_hybrid_2025} or SOH \citep{qiu_prompt-driven_2024} across multiple cathode chemistries (LFP, NMC) and temperature ranges, effectively treating different operating conditions as linguistic ``contexts'' rather than distinct statistical distributions. Taking a different route, Sun et al. \citep{sun_fine-tuning_2025} demonstrated the efficacy of dedicated Time-Series Foundation Models. By fine-tuning TimeGPT-1, they showed that a model pre-trained on general temporal dynamics could outperform specialized baselines for SOH estimation on a dataset of 143 distinct batteries, validating that temporal scaling laws hold true for electrochemical degradation.

By abstracting raw signals into a high-dimensional semantic space, these techniques decouple degradation logic from specific cell chemistries, paving the way for chemistry-agnostic foundation models adaptable through minimal prompt engineering or PEFT.

\subsubsection{Lifelong and Adaptive Learning}

A fundamental limitation of the conventional ``train-then-deploy'' methodology is that it produces a static model. In the real world, a battery is a dynamic system, which means its electrochemical characteristics evolve continuously due to aging, and its operational context can change, leading to a phenomenon known as concept drift, where the statistical properties of the data stream change over time.

To address this non-stationary nature of battery aging, researchers are leveraging the in-context learning abilities of LMs to create adaptive systems. Poh et al. \citep{poh_data-driven_2025} proposes a framework through distilling a large teacher model into a student that updates online. As the battery ages and the data distribution shifts, the model can adaptively learn from new, unlabeled data streams, ensuring it remains synchronized with the battery's evolving health state. Similarly, Feng et al. \citep{feng_adapting_2024} introduced a framework based on the concept of Test-Time Training (TTT). Unlike periodic updates, the TTT paradigm allows the model to perform continuous, incremental training using every single new data point as it is collected during the battery's operational lifetime.

By leveraging plasticity mechanisms such as in-context learning and efficient fine-tuning, these systems enable the model to co-evolve with the battery's degradation, addressing distribution shifts in long-term operation.

It is worth noting that federated learning (FL) has been actively explored for battery SOH estimation with conventional deep learning models, addressing privacy and data-sharing barriers across distributed fleets. However, the combination of FL with the LM paradigm, including federated pre-training or fine-tuning of foundation models across heterogeneous battery data sources, remains unexplored, representing an important direction discussed in \autoref{sec:roadmap_sec}.

\subsection{Integrating Domain Knowledge and Interpretability}

A well-documented criticism of conventional deep learning is its ``black box'' nature: outputs are not constrained by physical laws, potentially yielding predictions that are statistically plausible but electrochemically inconsistent. To address this, a growing body of research integrates domain knowledge, specifically electrochemical principles and physics-based models, directly into the LM framework to enhance robustness, physical consistency, and interpretability.

\subsubsection{Physics-Informed LMs}

The linguistic capabilities of LMs allows them to bridge the gap between qualitative physical laws and quantitative data. Li et al. \citep{li_coevolution_2025} utilize the LLM as a semantic orchestrator that translates natural language queries into executable simulation parameters, using established electrochemical models as ``tools'' to verify and ground its reasoning. Ren et al. \citep{ren_llm-enhanced_2025} take a more architectural approach by embedding physical constraints directly into the decoding stage. In their framework, the LM captures long-term dependencies, but the output generation is constrained by a decoder forced to adhere to valid degradation trajectories, ensuring that RUL predictions do not violate the monotonic nature of aging.

These contributions exemplify a neuro-symbolic evolution, using LMs as semantic bridges between physical laws and data-driven patterns. Notably, the framework of Li et al. \citep{li_coevolution_2025} closes a plan-execute-validate loop by invoking external electrochemical simulations to verify its own predictions, a capability with no counterpart in conventional BPHM.

\subsubsection{Knowledge-Enhanced and Multi-Modal Reasoning}

To combat hallucinations and enhance diagnostic logic, LMs are increasingly augmented with structured external knowledge. Ma et al. \citep{ma_knowledge-graph_nodate} demonstrate this with the FDRKG-LLM, which retrieves causal fault chains from a battery knowledge graph to guide the model's diagnosis. This ensures the model reasons via established failure mechanisms rather than statistical correlation. Similarly, BatteryGPT \citep{cheng_batterygpt_2024} employs multi-modal alignment to ground textual explanations in visual evidence, allowing the model to actually see a defect and explain it using accurate terminology derived from technical literature.

By retrieving structured knowledge and grounding explanations in visual and spectral evidence, these systems transition from statistical pattern matching to evidence-based causal reasoning, mitigating hallucination risks and enhancing industrial trustworthiness.

\subsection{Enabling System-Level Automation and Control}

The preceding section demonstrated how domain knowledge integration makes LM predictions more physically consistent and trustworthy. This trustworthiness, in turn, unlocks a further capability: deploying LMs not merely as predictors but as autonomous agents that act on their own outputs. Conventional methods typically produce a single numerical indicator such as SOH, lacking the contextual richness for direct integration into intelligent control systems. By combining reliable diagnostics with the reasoning and planning abilities unique to LMs, this gap can be effectively closed.

\subsubsection{LMs as Decision-Making Agents}

The reasoning capabilities of LMs allow them to function as high-level controllers that align battery operation with complex human intent. Yang et al. \citep{yang_llm-bas_2025} exemplify this with a dual-agent architecture  called LLM-BAS. Instead of simply outputting a health indicator, the system interprets the diagnostic state against dynamic electricity pricing and user preferences to formulate an optimal charging schedule. This demonstrates LMs' abilities in handling multi-objective optimization problems in zero-shot settings that could translate technical battery states into economic and operational decisions.

This direction redefines the BMS from a passive monitoring unit to a semantic decision-maker. By processing natural language constraints and multi-objective goals, LMs bridge the gap between human intent and low-level control logic, enabling greater autonomy and personalization in energy management.

\subsubsection{Automating the BPHM Research and Engineering Pipeline}

Beyond serving as predictive models, LMs are also deployed as automation tools for the BPHM research pipeline. From data preprocessing and feature engineering to model selection and hyperparameter optimization, LMs streamline traditionally manual, expertise-intensive tasks that bottleneck innovation.

Tuncel et al. \citep{tuncel_large_2025} pioneered this by using an LLM to orchestrate the entire pipeline for SOH prediction. Their framework guides through data preprocessing, feature ranking, and model tuning via structured prompts, effectively acting as an expert assistant that lowers the technical barrier for developing diagnostic models.

Beyond battery-specific pipelines, adjacent fields have demonstrated further automation potential. Wei et al. \citep{wei_efficient_2025} proposed a zero-shot framework for general time-series model selection, where the LM analyzes dataset characteristics and directly recommends the optimal algorithm and hyperparameters without any prior task-specific training. Zhang et al. \citep{zhang_enhanced_2025} similarly employed an LLM to autonomously optimize Wavelet Packet Transformation parameters for fault prediction in power equipment. Although neither study targets batteries directly, the underlying approach of using LLMs as automated model-selection and signal-processing agents is directly transferable to the BPHM pipeline.

By automating the technical intricacies of the modeling pipeline, these agents enable researchers and engineers to focus on high-level hypothesis generation rather than implementation details, potentially accelerating the discovery cycle for new diagnostic biomarkers and robust management algorithms.

\subsection{Comparative Summary of Representative Works}

To provide a more precise assessment of LM applications in BPHM, \autoref{tab:comparison} summarizes representative works across the four dimensions discussed above, comparing model type, target application, data scale, supervision level, approximate computational cost, deployment setting, and primary reported advantage.\footnote{Because none of the surveyed works report standardized metrics such as FLOPs or GPU-hours, Compute Cost is estimated from the training strategy: ``Low'' = prompt tuning or PEFT with frozen backbone; ``Moderate'' = full fine-tuning of a medium-scale model or one-time distillation; ``High'' = full LLM invoked at inference time. Deployment reflects the inference platform reported: ``GPU'' = workstation/server with no embedded deployment discussion; ``Edge-sim.'' = lightweight inference in a simulated or desktop environment; ``Edge'' = on-device inference on resource-constrained hardware explicitly targeted.}

Several observations emerge. First, the majority of current works operate at laboratory scale with GPU-server-based deployment. Second, pre-trained LMs consistently demonstrate advantages in data efficiency and transferability, particularly under low-data regimes \citep{bian_exploring_2024, qiu_prompt-driven_2024, peng_large-language-model-enabled_2025}. Third, a qualitatively different advantage lies in capabilities that conventional deep learning cannot offer: semantic orchestration of physics-based tools \citep{li_coevolution_2025}, physics-constrained decoding \citep{ren_llm-enhanced_2025}, and pipeline automation \citep{tuncel_large_2025}. However, controlled same-data-budget comparisons against well-tuned specialists remain absent, and evidence for deployment-ready edge inference is scarce. These remaining gaps motivate the systematic examination of challenges in the following section.

\begin{table}[pos=htbp]
\centering
\caption{A structured comparison of representative LM-based works in BPHM. Abbreviations: FM = Foundation Model, FT = Fine-Tuning, PEFT = Parameter-Efficient Fine-Tuning, SSL = Self-Supervised Learning, ZS = Zero-Shot, FS = Few-Shot.}
\label{tab:comparison}
\scriptsize
\begin{tabular}{p{2.2cm}p{1.5cm}p{1.8cm}p{1.2cm}p{1.3cm}p{1.3cm}p{1.2cm}p{1.8cm}}
\toprule
\textbf{Reference} & \textbf{Model Type} & \textbf{Application} & \textbf{Data Scale} & \textbf{Supervision} & \textbf{Compute Cost} & \textbf{Deploy.} & \textbf{Reported Advantages} \\
\midrule
\citet{fan_analyzing_2025} & LLM distill. & SOH/RUL & $\sim$100 cells & Supervised & Moderate & GPU & Data efficiency \\
\citet{peng_large-language-model-enabled_2025} & TS-FM (Lag-Llama) & SOH pred. & 143 cells & FT (0.72\%) & Low--Med. & GPU & Data efficiency \\
\citet{cheng_batterygpt_2024} & Multimodal LM & Anomaly det. & Lab-scale & FS/ZS & High & GPU & Transferability \\
\citet{lee_data-driven_2024} & LLM (parser) & Data extract. & 8000+ cells & Unsupervised & Moderate & GPU & Data scale \\
\citet{bian_exploring_2024} & LLM + prompt & SOC est. & Multi-chem. & PEFT & Low & GPU & Transferability \\
\citet{bian_hybrid_2025} & Hybrid LLM & SOC est. & Multi-chem. & PEFT & Low--Med. & GPU & Accuracy \\
\citet{qiu_prompt-driven_2024} & LLM + prompt & SOH est. & Multi-chem. & Prompt-FT & Low & GPU & Transferability \\
\citet{sun_fine-tuning_2025} & TS-FM (TimeGPT) & SOH est. & 143 cells & FT & Moderate & GPU & Accuracy \\
\citet{poh_data-driven_2025} & LM distill. & SOH (online) & Field-scale & Online & Low & Edge-sim. & Adaptability \\
\citet{feng_adapting_2024} & TTT-based LM & SOH est. & Lab + Field & SSL + TTT & Moderate & Edge & Robustness \\
\citet{li_coevolution_2025} & LLM (orchestr.) & Phys.-inform. & Lab-scale & Hybrid & High & GPU & Interpretability \\
\citet{ren_llm-enhanced_2025} & LLM + phys. dec. & RUL pred. & Lab-scale & Supervised & High & GPU & Phys. consist. \\
\citet{yang_llm-bas_2025} & LLM (agent) & Charging opt. & Simulation & ZS & High & GPU & Automation \\
\citet{tuncel_large_2025} & LLM (pipeline) & SOH pipeline & Lab-scale & Prompt & High & GPU & Automation \\
\bottomrule
\end{tabular}
\end{table}

\section{Current Challenges in Building LMs for BPHM}
\label{sec:challenges_sec}

\begin{figure}[pos=htbp]
    \centering
    \includegraphics[width=0.9\textwidth]{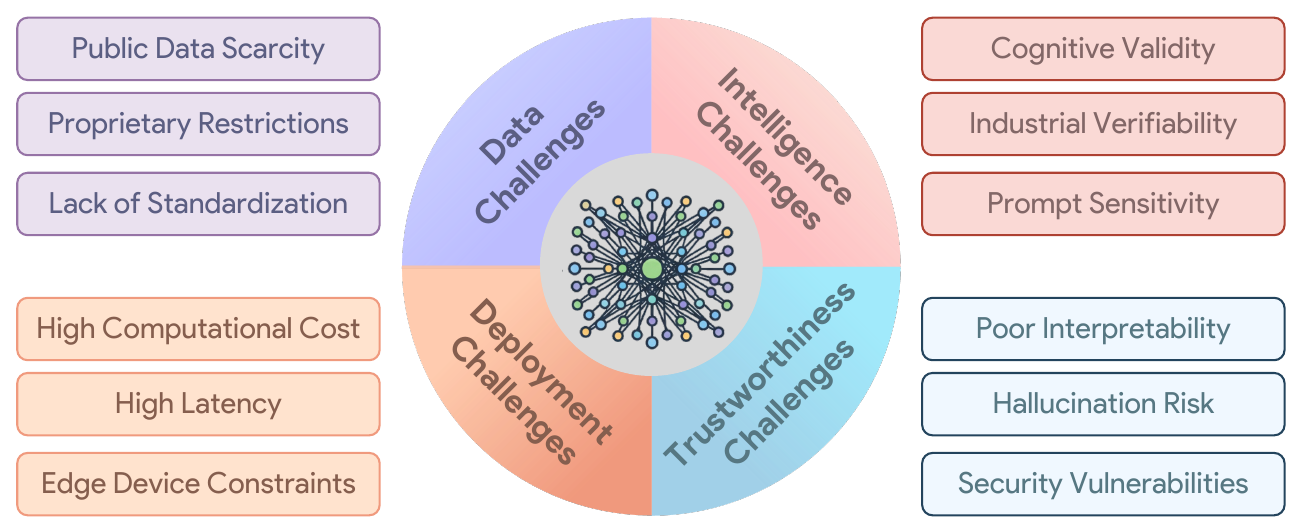}
    \caption{An overview of the key challenges in developing and deploying LMs for BPHM.}
    \label{fig:challenges}
\end{figure}

While the potential of LMs to revolutionize BPHM is substantial, the path from conceptual promise to industrial-grade deployment is fraught with significant and multifaceted challenges. These hurdles span the entire lifecycle of model development and application, from the foundational data layer to the final on-device implementation, as illustrated in \autoref{fig:challenges}.

\subsection{Paradigm-Level Limitations of LMs in BPHM}

The progress reviewed above demonstrates that the data-centric, pre-training--fine-tuning paradigm offers clear advantages over conventional task-centric approaches: it reduces labeled data requirements, enables cross-chemistry generalization, and unlocks qualitatively new capabilities such as emergent reasoning and agentic behavior. However, it is important to recognize that this paradigm shift does not eliminate the fundamental complexity of battery PHM. In practice, many downstream tasks still depend on task-specific labels (e.g., measured capacity for SOH, verified fault modes for diagnostics), application-specific protocols (e.g., reference performance test definitions that vary across manufacturers), domain-specific failure definitions (e.g., end-of-life thresholds that differ between EV and grid-storage applications), and deployment-specific constraints (e.g., latency, interpretability, and safety certification requirements). Pre-training on large corpora provides a richer initialization and can significantly reduce the labeled data requirement, but it does not eliminate the need for task-specific adaptation. In this sense, LMs partially alleviate task fragmentation but largely shift the bottleneck from model architecture design to data acquisition, curation, and adaptation engineering. Recognizing this nuance is essential for setting realistic expectations that LMs are best understood not as a universal solution that renders task-specific engineering obsolete, but as a powerful infrastructure that reduces the marginal cost of addressing each new battery task.

This paradigm-level limitation manifests across the four challenge areas examined below. On the data front, the shift toward pre-training amplifies the demand for large-scale, diverse, and high-quality corpora that are currently scarce and commercially sensitive. On the intelligence front, the emergent reasoning abilities observed in open-domain settings do not automatically transfer to physics-constrained battery problems, raising questions about cognitive validity. On the trustworthiness front, task-specific adaptation still requires domain-specific safety validation, interpretability, and robustness guarantees that generic pre-training cannot provide. On the deployment front, the computational overhead of LMs remains fundamentally at odds with the resource-constrained, latency-sensitive nature of embedded BMS hardware. The following subsections examine each of these challenge areas in detail.

\subsection{Data Challenges: Constructing Large-Scale Battery Datasets}

A fundamental challenge hindering the progress of LM-driven BPHM is the significant lack of public, large-scale, and diverse battery datasets. The successful pre-training of LM relies on a large and diverse corpus of data, but the field of battery science currently lacks resources comparable to those in natural language or computer vision domains. This data gap stems from a combination of structural and practical issues.

First, real-world battery operational data is often proprietary, held as a critical commercial asset. This severely restricts data flow to the research community, limiting collaborative model development. Publicly available data, while valuable, is often fragmented, small in scale, and unrepresentative of real-world diversity.

Second, even when data is accessible, quality and consistency remain problematic. Field data is heterogeneous, suffering from sensor noise, missing values, and unstandardized formats across manufacturers. Variations in sampling rates and logging protocols make cross-source harmonization exceedingly difficult, introducing noise and bias that compromise model reliability. Cleaning and aligning such disparate datasets is resource-intensive and constitutes a major bottleneck.

Third, even when collaborative frameworks such as federated learning are considered as a pathway to aggregate distributed data, significant practical barriers remain. Data ownership and intellectual property rights are ambiguous when multiple parties contribute to a jointly trained model: who owns the resulting model weights, and how are contributions fairly attributed? Standardization barriers are substantial, as different OEMs use incompatible BMS data formats, sampling rates, and sensor configurations, requiring non-trivial data harmonization layers before federated aggregation can proceed. Privacy concerns extend beyond raw data: even gradient updates shared during federated training can leak sensitive information through model inversion attacks \citep{kong_privacy-preserving_2024}. User-information leakage is a particularly acute concern for EV fleet data, where driving patterns and charging behavior can reveal personally identifiable information subject to regulations such as GDPR and China's Personal Information Protection Law (PIPL). Overcoming these barriers will require not only technical solutions such as differential privacy and secure aggregation, but also governance frameworks, data-use agreements, and regulatory clarity.

\subsection{Intelligence Challenges: Cognitive Validity and Industrial Verifiability}
While LMs have demonstrated remarkable emergent abilities and strong generalization in open-domain tasks, their existence and effectiveness in industrial scientific domains such as BPHM have not yet been systematically validated. Currently, the understanding of the intelligence augmentation provided by large language models is still primarily based on empirical observation, lacking a solid theoretical foundation and reproducible evidence under controlled industrial conditions.

On one hand, the reasoning and self-supervised learning abilities that LMs exhibit in open-domain tasks may not necessarily transfer to the highly constrained, data-scarce, and physics-dominated environment of battery operation. In such applications, a successful model requires more than just pattern recognition; it demands an understanding of causal relationships, physical consistency, and the reliability of its own reasoning. While increasing the parameter scale of an LM demonstrably enhances its capacity for memorization and representation, it does not guarantee a corresponding improvement in performance on tasks that are governed by strict physical laws. The assumption that scale promotes intelligence has been shown to be effective in language models, but remains an unverified and potentially flawed assumption in the context of scientific and engineering problems.

On the other hand, the intelligent performance of LMs is highly dependent on the design of prompts and the construction of the input context. It has been observed that different phrasing or variations in context length can lead to dramatically different outputs, suggesting that the model's apparent intelligence is more a result of statistical fitting to external conditions rather than a genuine cognitive understanding. This non-robust and often irreproducible intelligent behavior introduces significant uncertainty when applied to industrial BPHM tasks. Furthermore, current LM capabilities such as Chain-of-Thought (CoT) reasoning \citep{wei_chain--thought_2022} and agentic capabilities are based on linguistic formalism rather than physically verifiable logic. This inherent limitation further constrains their reliability in scientific reasoning tasks like the prediction of battery degradation, where the underlying causal mechanisms are of great importance.

It is also important to note that stronger generalization does not automatically translate into superiority over existing approaches. In many practical settings, a carefully engineered task-specific model, such as a physics-informed neural network calibrated for a single cell chemistry or an LSTM trained on a well-characterized degradation dataset, can still outperform a generalist LM on that specific task. Battery PHM is caught between two forces that pull in opposite directions. On one side, the field is extremely fragmented: dozens of chemistries, operating conditions, and degradation modes make it impractical to build and maintain a dedicated specialist for every combination, and this fragmentation is precisely where LMs should excel. On the other side, many BPHM tasks are safety-critical, and for any single high-stakes function, a mediocre generalist is worse than a strong specialist. What remains unknown is where the crossover lies: at what level of task diversity and data scarcity does a pre-trained LM begin to outperform the specialist alternative on a total cost-of-ownership basis? The field currently lacks the standardized, head-to-head benchmarks needed to answer this question. Until such benchmarks exist, the more productive path forward may be to treat LMs not as replacements for specialists but as shared pre-trained backbones from which lightweight, task-specific heads can be derived, an approach that has already proven effective in NLP \citep{qiuPretrainedModelsNatural2020} and computer vision \citep{goldblumBattleBackbonesLargeScale2023} but remains largely unexplored in battery PHM.

\subsection{Trustworthiness Challenges: Interpretability, Security, Safety, and Robustness}
\label{sec:trustworthiness}
Real-world applications of LMs require a high level of trustworthiness. While LMs offer significant advancements in parameter scale, semantic understanding, and task generalization compared to conventional DL models, they also introduce a new and more complex set of challenges related to interpretability, security, safety, and robustness.

First, the internal representations of LMs are extraordinarily complex, rendering their decision-making logic opaque and difficult to map onto understandable physical or chemical mechanisms. Standard LMs are purely statistical and have no inherent understanding of electrochemistry, making them classic black box systems. In safety-critical scenarios, the inability of a model to clearly articulate the basis for its predictions makes it exceptionally difficult for engineers and regulatory bodies to establish trust.

More alarmingly, LMs are prone to inherent issues of ``hallucination,'' where they can generate outputs that are plausible on the surface but are devoid of any physical basis \citep{huang_survey_2025}. For example, when tasked with explaining battery capacity fade, an LM might spontaneously invent a non-existent aging mechanism or misapply an established experimental law. Such errors are particularly dangerous in scientific research and industrial diagnostics because their outputs possess a high degree of superficial credibility while being fundamentally incorrect.

Furthermore, LMs introduce security threats with direct physical consequences in battery applications. Data poisoning---for instance, a compromised charging station injecting biased voltage readings into a fleet-wide training pipeline---can cause the global model to systematically underestimate degradation rates, with effects that may not manifest until months of accumulated bias lead to unexpected field failures \citep{alberMedicalLargeLanguage2025a}. Adversarial perturbations applied to impedance spectra or voltage profiles could trigger false thermal runaway predictions or, conversely, suppress genuine safety warnings \citep{wang_adversarial_2022, shayegani_survey_2023}. Prompt injection attacks can manipulate the model into generating unsafe control strategies \citep{liuPromptInjectionAttack2025}. In federated settings, model inversion attacks on shared gradient updates could reconstruct proprietary cell chemistry information \citep{kong_privacy-preserving_2024}. Concept drift, where the statistical properties of battery data shift over time due to aging, further compounds these challenges, as a model's performance can degrade silently after deployment. These vulnerabilities underscore that in battery PHM, cybersecurity threats have direct implications for physical safety, commercial integrity, and regulatory compliance.

\subsection{Deployment Challenges: Efficiency and Responsiveness}

The enormous parameter scale and high computational complexity of LMs present formidable challenges for their deployment on resource-constrained platforms such as on-vehicle BMS or other edge devices. There is a fundamental mismatch existing between the computational requirements of LMs and the capabilities of the hardware typically found in embedded systems. Whereas cloud servers can provide hundreds of gigabytes of GPU memory and high-bandwidth interconnects to support LM inference, BMS chips are severely limited in terms of power consumption, memory, and processing power. This discrepancy creates a fundamental trade-off between the performance of the model and the energy efficiency of the system.

In addition, the real-time responsiveness of LMs often fails to meet the stringent latency requirements of BPHM tasks. In dynamic scenarios, such as fast charging, the onset of thermal runaway, or the sudden occurrence of a fault, the BMS should be capable of making predictions and decisions within a millisecond-level timeframe. However, the inference process for LMs is typically associated with higher latency and significant bandwidth consumption, particularly if it relies on cloud-based APIs. This makes their direct application in embedded systems, where low-latency, real-time control is critical, a significant engineering and scientific challenge. The central problem, therefore, is how to effectively deploy LMs to the edge without sacrificing their intelligent performance.

\section{Solutions and Future Roadmap}
\label{sec:roadmap_sec}

\begin{figure}[pos=htbp]
    \centering
    \includegraphics[width=0.6\textwidth]{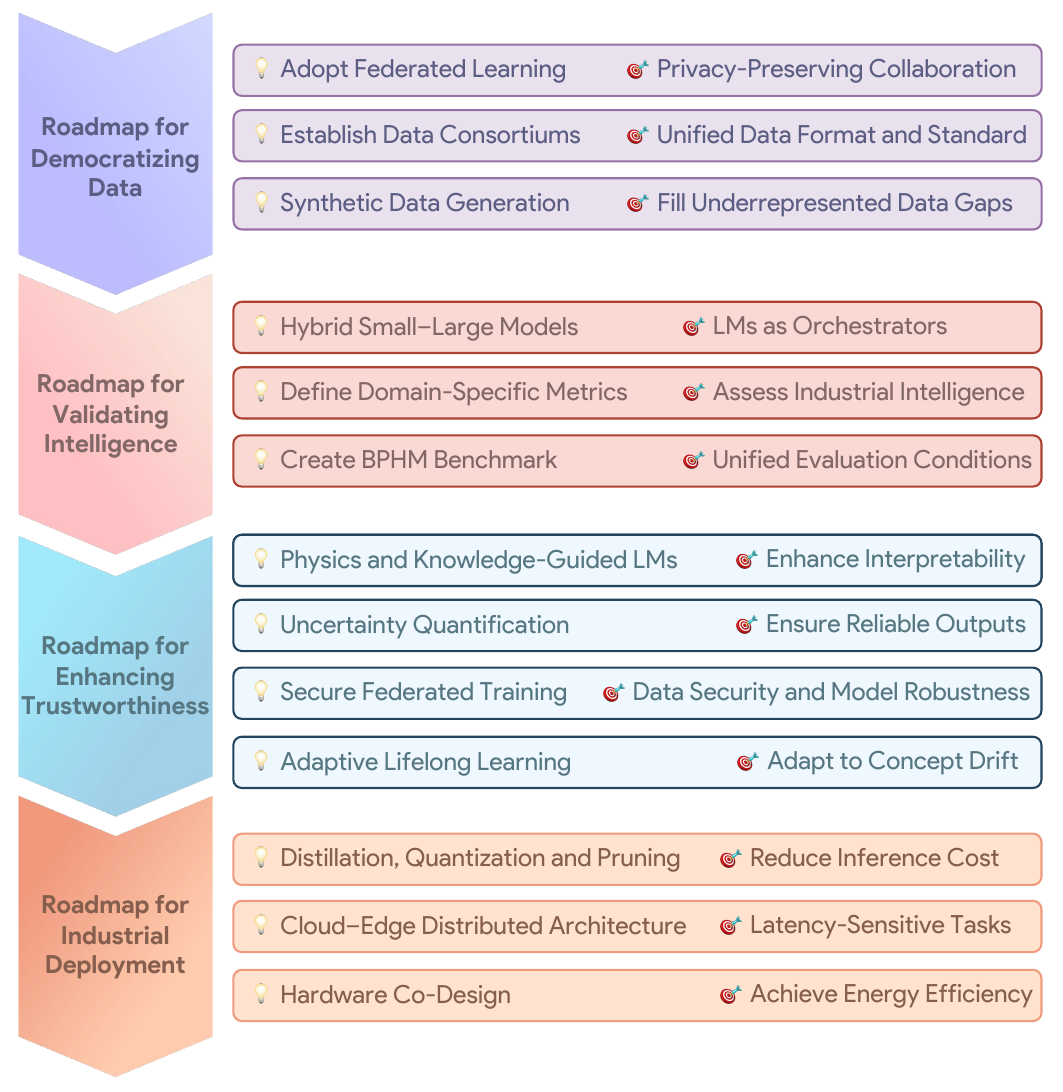}
    \caption{A proposed future roadmap for advancing LM-driven BPHM. The roadmap is structured around four key pillars and specific strategic initiatives and their target objectives are outlined for each pillar.}
    \label{fig:roadmap}
\end{figure}

To address the multifaceted challenges outlined in the previous section, a concentrated and strategic research effort is required. This section proposes a systematic future roadmap organized around four core challenge areas: data accessibility, model intelligence, trustworthiness, and deployment. For each area, we outline key strategic directions and specific technical pathways that can collectively drive the LM-driven BPHM field from its current nascent stage to robust, industrial-scale applications. A visual summary of this proposed roadmap is presented in \autoref{fig:roadmap}.

\subsection{Democratizing Data: Building Open and Scalable Battery Data Ecosystems}

Addressing data scarcity in BPHM requires a shift from strategies relying on fragmented, proprietary datasets to a more open, collaborative, and technology-driven data ecosystem. Critically, building meaningful battery foundation models will require data that goes far beyond the controlled laboratory datasets currently available. Field data from real-world deployments---EV fleet telemetry capturing diverse driving patterns and climate conditions, stationary energy storage systems operating under grid-level cycling regimes, and charging infrastructure logs recording thousands of daily sessions across heterogeneous vehicle populations---are essential for capturing the full distributional diversity of battery degradation in practice. However, accessing such data at scale faces formidable practical barriers that must be explicitly acknowledged and addressed.

\subsubsection{Enabling Privacy-Preserving Data Collaboration Through Federated Learning}

A key technological solution is the adoption of Federated Learning (FL). FL provides a privacy-preserving framework for collaborative model training, allowing multiple organizations to jointly train a powerful, global model without ever sharing their raw, sensitive data \citep{mcmahan_communication-efficient_2023}. This approach directly addresses the commercial secrecy and data ownership concerns, enabling the aggregation of knowledge from diverse, distributed datasets to build more robust and generalizable LMs. In concrete deployment scenarios for battery PHM, FL could enable Original Equipment Manufacturers (OEMs) to collaboratively train a shared degradation model across their respective vehicle fleets without exposing proprietary cell chemistry details or customer usage patterns. Similarly, charging network operators could contribute aggregated session-level degradation signals to a federated model while maintaining commercial confidentiality over station utilization data.

To address the practical barriers, several technical safeguards should be integrated into FL pipelines for battery PHM. Differential privacy mechanisms can bound the information leakage from shared gradient updates, mitigating the model inversion risks that threaten proprietary chemistry data and user privacy \citep{kong_privacy-preserving_2024}. Secure aggregation protocols ensure that the central server only observes the aggregated model update rather than individual contributions, further protecting commercial confidentiality. To overcome the standardization barriers arising from incompatible BMS data formats across OEMs, federated architectures should incorporate preprocessing harmonization layers that normalize sampling rates, sensor configurations, and feature definitions before local training begins. Finally, contribution-weighted aggregation strategies can fairly attribute value to each participant's data, incentivizing high-quality contributions and providing a basis for equitable IP governance.

\subsubsection{Establishing Pre-Competitive Industrial Data Consortium}

Concurrently, a coordinated effort to establish a pre-competitive industrial data consortium is essential. Precedents in adjacent domains demonstrate the feasibility of this approach: the Materials Genome Initiative \citep{MaterialsGenomeInitiative2013} has accelerated materials discovery through shared computational and experimental databases, while the autonomous driving community has benefited from large-scale open datasets such as the Waymo Open Dataset \citep{sunScalabilityPerceptionAutonomous2020} and nuScenes \citep{caesarNuScenesMultimodalDataset2020} that enabled rapid benchmarking across research groups. A battery data consortium, driven by collaboration among academia, industry, and government agencies, should prioritize three technical foundations: (1) a unified cell-level data schema that standardizes how voltage, current, temperature, and impedance measurements are recorded, timestamped, and annotated with metadata (cell chemistry, manufacturer, cycling protocol, ambient conditions); (2) provenance tracking that documents the full data lineage from raw sensor acquisition through preprocessing, enabling reproducibility; and (3) tiered access control that distinguishes between fully open benchmark subsets for academic research and restricted industrial subsets available under data-use agreements. Given that the success of many leading language models relies on access to large-scale proprietary datasets, the emerging risk of data monopolies in battery PHM cannot be ignored. Therefore, the future development roadmap should emphasize the democratization of large-scale data resources through continuous public-private collaboration, ensuring a fair, transparent, and competitive ecosystem.

\subsubsection{Enriching Datasets Through Synthetic Generation and Foundation Model Transfer}
\label{sec:synthetic}
To further enrich available datasets, the roadmap should also include the systematic application of advanced data augmentation techniques. The potential of generative models, such as diffusion-based frameworks like DiffBatt \citep{eivazi_diffbatt_2024}, should be fully exploited to synthesize high-fidelity, realistic battery data. This synthetic data can be used to fill critical gaps in real-world datasets, for example, by generating data for rare fault modes or under-represented operating conditions, thereby enhancing the generalization and robustness of the trained models. Battery digital twins (DTs)---high-fidelity virtual replicas continuously synchronized with real-time operational data---offer a complementary route: by running physics-based simulations (e.g., electrochemical-thermal coupled models) across a wide parameter space of chemistries, temperatures, C-rates, and degradation modes, DTs can produce large volumes of physically consistent training data. This is particularly valuable for generating data for rare but critical scenarios, such as internal short circuits or extreme temperature excursions, that are dangerous or impractical to reproduce experimentally but essential for training robust LMs. Finally, a crucial direction is to reduce the dependency on battery-specific data altogether. By leveraging the vast and publicly available corpora of general time-series data, it is possible to pre-train time-series foundation models. These models can then be fine-tuned on much smaller, battery-specific datasets, inheriting the powerful generalization capabilities learned from the broader data landscape and reducing the burden of domain-specific data collection.

\subsection{Validating Intelligence: Assessing LM Capabilities in BPHM Settings}

To move beyond the current empirical understanding of LM capabilities in BPHM, the future roadmap should focus on developing a rigorous, systematic framework for validating their cognitive effectiveness in an industrial scientific context.

\subsubsection{Designing Hybrid Architectures with Smaller Models}

A shift toward Small Model Augmentation (SMA) strategies is necessary. Instead of relying on ever-larger monolithic models, the development of hybrid intelligent architectures should be prioritized. In this paradigm, the LM functions as a high-level knowledge memory and task orchestrator, while lightweight specialized models or traditional physics-based models handle fine-grained reasoning and numerical computations. This workflow resembles an agent-based system, where LMs are able to call certain tools for different purposes. Such structured modularity reduces dependence on model scale and enhances both the efficiency and reliability of system-level intelligence.

\subsubsection{Developing Domain-Specific Intelligence Metrics}
Standard prediction metrics such as RMSE and MAE evaluate ``what'' a model outputs but reveal nothing about ``why''. A model that achieves low RMSE on SOH prediction may have learned the true relationship between impedance growth and lithium inventory loss, or it may have simply memorized a degradation curve shape from the training distribution. These two models will be indistinguishable on in-distribution test sets yet diverge catastrophically under novel conditions. Resolving this ambiguity requires a hierarchical validation protocol that probes progressively deeper levels of electrochemical understanding: (1)~\textit{Physical plausibility}, verifying that predictions obey thermodynamic and kinetic constraints (e.g., monotonic resistance growth, thermodynamically valid OCV curves), with violations detected automatically against physics-based simulations or digital twin references; (2)~\textit{Calibration quality}, assessing whether predicted confidence aligns with empirical accuracy via metrics such as Expected Calibration Error (ECE), which is especially critical as the battery's operational envelope evolves with aging and training coverage diminishes in late-life regimes; (3)~\textit{Diagnostic sensitivity}, evaluating whether the model can detect subtle degradation signatures (lithium plating onset, micro-short circuits, electrolyte decomposition) hundreds of cycles before macroscopic failure, using curated fault-injection datasets with detection lead time and false positive rate as key metrics; and (4)~\textit{Generalization under distributional shift}, constructing held-out splits along physically meaningful axes (e.g., train on NMC / test on LFP; train at 25\textdegree{}C / test at 45\textdegree{}C) rather than random splits, and stress testing under abnormal conditions to probe whether the model degrades gracefully or produces dangerously overconfident predictions.

\subsubsection{Creating Standardized Benchmarks}
Finally, the creation of an Industrial Intelligence Benchmark for BPHM is necessary. This benchmark should be a standardized testing platform that encompasses a wide range of battery chemistries, degradation pathways, and noise levels. By providing unified evaluation tasks, prompt templates, and performance metrics, this platform will provide a standardized and reproducible basis for validating the effectiveness of LMs in the industrial intelligence domain, enabling fair and meaningful comparisons across different models and methodologies.

\subsection{Enhancing Trustworthiness: Physics-Guided, Secure, Safe, and Adaptive LMs}

To overcome the significant trustworthiness challenges of LMs in BPHM, the future roadmap should pivot from generic solutions towards a multi-layered, domain-specific defense strategy.

\subsubsection{Integrating Physical Principles and Causal Inference for Interpretability}
Methods for enhancing interpretability should go beyond tracing model decisions and instead directly embed physical principles into the architecture. This shifts the approach from opaque correlation toward causal diagnostics. Key technical directions include the integration of Physics Informed Neural Networks, where the partial differential equations of electrochemical models are incorporated into the loss function so that predictions follow fundamental physical laws \citep{wang_physics-informed_2025, wen_physics-informed_2024}. In addition, Knowledge Graph guided reasoning can provide structured and verifiable prior knowledge about material properties and degradation mechanisms, which constrains the model's inference pathways. The use of causal inference tools is also important for moving from correlation to causation, allowing the model to identify the underlying reasons for state of health degradation. Battery digital twins, introduced in \autoref{sec:synthetic} as a data-generation tool, can also serve a complementary role here as real-time physics oracles: the LM proposes hypotheses or predictions based on learned patterns, and the digital twin validates them against established electrochemical models, rejecting outputs that violate physical constraints. As the physical battery ages and its characteristics drift, the digital twin can be continuously re-calibrated using incoming sensor data, providing an up-to-date reference against which LM predictions are benchmarked and corrected. While DTs ground LMs in physics, LMs can automate the traditionally laborious processes of DT parameterization and calibration, creating a self-improving battery intelligence ecosystem in which physics-based simulation and data-driven learning reinforce each other continuously.

\subsubsection{Quantifying and Constraining Uncertainty}
Recognizing that hallucination is an inherent limitation of LMs, the roadmap should focus on actively mitigating this risk. This requires a layered framework that quantifies uncertainty and enforces strict guardrails. A key step is the adoption of probabilistic modeling and uncertainty quantification, where model outputs are represented as distributions rather than single values. This can be implemented through Bayesian approaches or by using probabilistic generative models. Moreover, these methods can be combined with Retrieval Augmented Generation to ground factual statements in verifiable data and applies constrained generation techniques to prevent the model from producing physically implausible results.

\subsubsection{Ensuring Data Security and Model Robustness in Federated Environments}
Addressing the multifaceted security threats identified in \autoref{sec:trustworthiness} requires a defense-in-depth architecture rather than any single countermeasure. At the data ingestion stage, incoming telemetry streams should be validated before they enter the training pipeline. Statistical process control methods and autoencoder-based anomaly detectors \citep{sun_anomaly_2023} can monitor raw voltage, current, and temperature signals in real time, flagging signatures of sensor tampering or systematic bias from compromised charging infrastructure. Battery digital twins add a complementary, physics-based check: by running electrochemical-thermal simulations alongside incoming field data, a DT can catch measurements that look statistically normal but are physically inconsistent, for instance voltage profiles that appear smooth yet violate the expected relationship between state of charge and open-circuit voltage for a given chemistry. Purely statistical filters would miss this kind of carefully crafted poisoning.

Securing the federated training process requires several coordinated mechanisms. Differential privacy protocols \citep{kong_privacy-preserving_2024} should bound the information leakage per gradient update through calibrated noise injection, protecting proprietary cell chemistry data and user behavioral patterns from model inversion attacks. The aggregation step itself also needs hardening: robust rules such as coordinate-wise median or trimmed mean \citep{yin_byzantine-resilient_2018} can replace naive averaging, allowing the central server to down-weight or exclude outlier updates from compromised participants. Data provenance tracking, in which each participant's contribution is logged with cryptographic audit trails, further supports retrospective identification of poisoned update sources.

The model itself should undergo BPHM-specific adversarial training with perturbations drawn from domain-relevant noise profiles (sensor drift, quantization error, temperature-dependent measurement bias, and impedance spectroscopy artifacts) rather than generic perturbations \citep{wang_adversarial_2022}. Training against these realistic distributions produces representations more robust to the noise actually encountered in battery data. On the output side, physics-constrained guardrails should reject predictions that violate electrochemical laws, such as non-monotonic internal resistance under normal aging or thermodynamically invalid open-circuit voltage curves. With defenses at every stage from sensor to output, no single point of compromise can propagate unchecked through the pipeline.

\subsubsection{Enabling Adaptive Lifelong Learning}
A trustworthy model should adapt to concept drift as a battery ages. The roadmap should therefore embrace the paradigm of Test-Time Training, or online adaptation. In this framework, the deployed model continuously fine-tunes its parameters using every new data point it receives. To make this computationally feasible on resource-constrained BMS hardware, efficient on-device PEFT strategies are essential. This enables personalized lifelong learning with minimal computational overhead.

\subsection{Enabling Industrial Deployment: Efficient and Scalable Model Integration}

Translating LM capabilities into industrial BPHM practice requires addressing both the computational overhead of large-scale models and the stringent latency, interpretability, and safety requirements of embedded battery management systems.

\subsubsection{Compressing LMs Through Distillation, Quantization and Pruning}

Deploying LM-enabled functions such as real-time SOH/SOC estimation, adaptive charging optimization, and online anomaly detection on edge hardware requires transferring LM capabilities into sufficiently compact models. A primary pathway is the development of a hierarchical knowledge distillation and multi-scale model architecture. This requires building a layered distillation pipeline in which the full knowledge of a large cloud-based teacher model is transferred step by step to a set of smaller student models designed for edge deployment. In parallel, model compression techniques such as quantization and pruning should be applied to further reduce computational and memory overhead. Quantization maps high-precision parameters to lower-bit representations to accelerate on-device inference, while pruning removes redundant weights or neurons to streamline the model architecture. Together, these strategies enable lightweight, resource-efficient student models suitable for real-time edge deployment.

\subsubsection{Designing Distributed Architectures for Cloud-Edge-Device Deployment}
Rather than compressing a single monolithic model, a more scalable strategy is to distribute LM-enabled functions across a hierarchical cloud-edge-device architecture, assigning each function to the computational tier that matches its latency tolerance, interpretability requirements, and current technological maturity. \autoref{tab:deployment} summarizes this tiered deployment framework.

\textbf{Tier 1 (GPU-server analytics)} encompasses functions that tolerate seconds-to-minutes latency and can leverage the full computational power of server-class GPUs. These include fleet-level degradation modeling, what-if simulation for battery second-life assessment, long-term lifetime prediction, and offline anomaly root-cause analysis. Such functions are realistically deployable in the short term using existing LM infrastructure and represent the most immediate value proposition of LMs in battery PHM.

\textbf{Tier 2 (edge-distilled inference)} covers functions requiring millisecond-level latency that can be addressed by lightweight models distilled from larger GPU-server-based teachers via the compression techniques described in the preceding subsubsection. These include real-time SOH and SOC estimation during normal operation, adaptive charging optimization, and online anomaly detection. Achieving this tier requires successful knowledge distillation and model compression, making it a medium-term target.

\textbf{Tier 3 (not yet feasible)} contains safety-critical functions that currently cannot tolerate the latency, opacity, or reliability characteristics of LM-based systems. These include real-time thermal runaway prevention, sub-millisecond fault response during fast charging, and safety-critical control actions where incorrect decisions have immediate physical consequences. For these functions, the lack of real-time guarantees, the opacity of LM decision-making, and the absence of formal verification methods make direct LM deployment premature. Hybrid approaches, where LMs inform but do not directly execute safety-critical control, may offer a transitional pathway.

\begin{table}[pos=htbp]
\centering
\caption{A tiered deployment readiness framework for LM-enabled battery PHM functions.}
\label{tab:deployment}
\scriptsize
\begin{tabular}{p{3.2cm}p{1.0cm}p{1.3cm}p{1.5cm}p{1.4cm}p{2.5cm}}
\toprule
\textbf{Function} & \textbf{Tier} & \textbf{Latency Tol.} & \textbf{Interp. Req.} & \textbf{Maturity} & \textbf{Key Barrier} \\
\midrule
Fleet degradation modeling & 1 & min & Low & High & Data access \\
What-if simulation & 1 & min & Low & Medium & Physics grounding \\
Lifetime prediction & 1 & sec--min & Medium & Medium & Cross-chem. valid. \\
Offline anomaly analysis & 1 & min & Medium & Medium & Labeled fault data \\
\addlinespace
Real-time SOH/SOC est. & 2 & ms & Medium & Low--Med. & Model compression \\
Charging optimization & 2 & sec & Medium & Low & Safety certification \\
Online anomaly detection & 2 & ms--sec & High & Low & False positive rate \\
\addlinespace
Thermal runaway prev. & 3 & sub-ms & Critical & Very low & Real-time guarantee \\
Fast-charge fault resp. & 3 & sub-ms & Critical & Very low & Formal verification \\
Safety-critical control & 3 & sub-ms & Critical & Very low & Interpretability, cert. \\
\bottomrule
\end{tabular}
\end{table}

Realizing this hierarchical architecture also requires decomposing the LM framework into modular components that can be intelligently distributed across the computational hierarchy. Global inference and long-term learning are executed on GPU servers, while latency-sensitive, real-time predictions are handled at the edge nodes. Furthermore, research into dynamic reasoning mechanisms, which activate only a subset of model parameters based on the complexity of the input data, can significantly reduce average inference costs and enable on-demand computation.

\subsubsection{Co-Designing Next-Generation Hardware and Energy-Efficient Algorithms}
The Tier 3 functions identified in \autoref{tab:deployment}, currently infeasible due to sub-millisecond latency and formal verification requirements, will ultimately depend on hardware breakthroughs. Looking toward these future deployment needs, the roadmap should include a focus on the co-design of novel, energy-efficient hardware and tailored algorithms. Emerging architectures such as neuromorphic computing, brain-inspired chips, and photonic AI promise orders-of-magnitude improvements in power efficiency. The development of new model architectures specifically designed to leverage the unique characteristics of these hardware platforms, such as event-driven Spiking LMs \citep{zhu_spikegpt_2024}, could provide a disruptive solution for ultra-low-power deployment at the edge, ultimately enabling the full integration of advanced AI into the next generation of battery management systems.

\section{Conclusions}
\label{sec:conclusions}

This paper has provided the first systematic review of LM applications in BPHM. We began by elucidating the core enabling technologies, including the Transformer architecture, self-supervised learning, multimodal data fusion, and parameter-efficient fine-tuning, and then synthesized the rapid progress being made along four dimensions: mitigating data scarcity, enhancing generalization, integrating domain knowledge for interpretability, and enabling system-level automation. Our comparative analysis reveals that LMs offer their clearest advantages in low-data, cross-domain settings, while also unlocking qualitatively new capabilities such as physics-constrained reasoning and agentic decision-making.

However, this paradigm shift does not eliminate the fundamental complexity of battery PHM. LMs are best understood not as universal replacements for task-specific models, but as a powerful shared infrastructure that reduces the marginal cost of addressing each new battery task while still requiring domain-specific adaptation, safety validation, and interpretability guarantees. Critical challenges remain across data accessibility, cognitive validity, trustworthiness, and deployment feasibility.

The future roadmap proposed in this review calls for a unified effort to build collaborative data ecosystems through federated learning and pre-competitive consortia, validate LM intelligence through hierarchical domain-specific metrics, embed physical principles to ensure trustworthiness, and develop efficient cloud-edge-device deployment architectures. By following this roadmap, researchers and practitioners can advance battery management from a set of isolated predictive tasks toward a holistic, autonomous system capable of reasoning, adaptation, and control---paving the way for next-generation battery systems that are safer, longer-lasting, and more reliable.

\bibliographystyle{cas-model2-names}
\bibliography{references}

\end{document}